\documentclass[10pt,twocolumn,a4paper]{article} % TWO

\usepackage{graphicx}
\usepackage{setspace}

\usepackage[square,numbers]{natbib}

\usepackage{graphics} % for pdf, bitmapped graphics files
\usepackage{epsfig} % for postscript graphics files
\usepackage{textcomp}
\usepackage[T1]{fontenc}

\usepackage[hidelinks]{hyperref}

\usepackage{amsmath,amsfonts,amssymb}

\usepackage{bm}

\usepackage{subcaption}

\usepackage{authblk}

\title{Localizing Multiple Radiation Sources Actively with a Particle Filter}

\author[1]{Tomas Lazna}
\author[1]{Ludek Zalud}
\affil[1]{Central European Institute of Technology, Brno University of Technology, Brno, Czech Republic\\ 
	\textit{\{tomas.lazna,ludek.zalud\}@ceitec.vutbr.cz}}

\begin{document}

\maketitle

%%%%%%%%%%%%%%%%%%%%%%%%%%%%%%%%%%%%%%%%%%%%%%%%%%%%%%%%%%%%%%%%%%%%%%%%%%%%%%%%
\begin{abstract}

We discuss the localization of radiation sources whose number and other relevant parameters are not known in advance. The data collection is ensured by an autonomous mobile robot that performs a survey in a defined region of interest populated with static obstacles. The measurement trajectory is information-driven rather than pre-planned, and the localization exploits a regularized particle filter estimating the sources' parameters continuously. Regarding the dynamic robot control, this switches between two modes, one attempting to minimize the Shannon entropy and the other aiming to reduce the variance of expected measurements in unexplored parts of the target area; both of the modes maintain safe clearance from the obstacles. The performance of the algorithms was tested in a simulation study based on real-world data acquired previously from three radiation sources exhibiting various activities. Our approach reduces the time necessary to explore the region and to find the sources by approximately 40~\%; at present, however, the method is unable to reliably localize sources that have a relatively low intensity. In this context, additional research has been planned to increase the credibility and robustness of the procedure and to improve the robotic platform autonomy.

\end{abstract}

%%%%%%%%%%%%%%%%%%%%%%%%%%%%%%%%%%%%%%%%%%%%%%%%%%%%%%%%%%%%%%%%%%%%%%%%%%%%%%%%
\section{Introduction}

Radiation detection and the localization of radioactive sources are tasks of major significance in many fields and facilities, including nuclear power plants, environmental monitoring, and search for uncontrolled sources. Traditionally, these procedures have relied on human operatives, who are required to enter potentially contaminated areas; the advancement in robotics has nevertheless allowed radiation mapping with unmanned vehicles. The robots can survey hazardous environment, eliminating nuclear risk to human health, and are capable of navigating through complex terrain to locate and identify radiation sources. By using autonomous vehicles, we can also increase the efficiency and accuracy of the process. This article has been designed to present an approach for localizing multiple radiation sources without prior knowledge of their number and other relevant parameters. We propose a method that localizes sources by using a particle filter combined with an active planning strategy, increasing the task performance efficiency. 

The problem of finding sources of ionizing radiation via robotic assets has been thoroughly studied in the literature. An overview of different approaches to active localization (i.e., the measurement trajectory is not pre-planned or controlled by a human operator), including their comparative analysis, is offered in \cite{mcgill_robot_2011}. Several techniques rely on pixel detectors or Compton cameras, which provide various degrees of directional information. The set of articles on mapping or passive localization comprises, for instance, reference \cite{kim_3d_2017}, where a 3D radiation image is reconstructed to enable locating a single source. The authors of \cite{stibinger_localization_2020} introduce a comprehensive simulation tool for Timepix detectors, verifying their instrument via using a micro aerial vehicle to retrieve a source. An additive point source localization algorithm is presented in \cite{vavrek_reconstructing_2020}, demonstrating its ability to find up to four radionuclides by means of a custom handheld device. An active localization method utilizing a Compton imaging device and the maximum likelihood-expectation maximization algorithm is described in \cite{lee_active_2018}; the authors exploit a Fisher information matrix (FIM) to identify an optimal sequence of dwell points. The central deficiencies of gamma cameras, namely, the long acquisition time and poor angular resolution, are addressed in \cite{ardiny_autonomous_2019}; here, an optimal data acquisition strategy to suit the camera's parameters is outlined through multi-criteria decision-making, delivering better results than the behavior-based approach. Article \cite{kishimoto_path_2021} applies principal component analysis to previous measurements to determine the direction of the next dwell point, the localization relying on a simple back-projection; there is the possibility of locating multiple sources, which nevertheless requires an input from a human operator. 

Other articles discuss common omnidirectional detectors; at this point, we can focus on those that investigate passive localization. A method for extracting directional information from an acquired dataset and finding intersections via maximum likelihood estimation is presented in \cite{tan_fast_2022}. The authors of \cite{bird_robot_2019} then propose a platform based on the Robot Operating System (ROS) to systematically map an indoor environment in which radiation hotspots are definable. The approach characterized in \cite{chin_efficient_2011} relies on a static network of detectors and presents a hybrid particle filter supported by a mean-shift algorithm capable of locating an unknown number of sources.

An associated cluster of articles embraces active localization; here, studies considering a single source are referred to first. The procedure set out in \cite{huo_autonomous_2020} localizes the source via a particle filter enhanced with a Markov chain Monte Carlo method; the search strategy alone adopts a partially observable Markov decision process subsuming a reward function based on the Shannon entropy. Another concept that employs the entropy is exposed in \cite{pinkam_informative_2020}. Further, article \cite{liu_localizing_2022} proposes a combination of a particle filter and an unscented Kalman filter to estimate the source position in each axis separately; the robot is driven directly towards the point where the source is anticipated. 

Finally, related work on the active localization of multiple sources is summarized. Article \cite{mascarich_autonomous_2022} focuses on sophisticated radiation mapping rather than source localization; the proposed framework is able to reconstruct a 3D map with an unmanned aircraft system (UAS) in a satellite navigation-denied environment, and different isotopes can be distinguished. Another UAS-based approach exploits contour following supported by sampling in a suitable region of interest \cite{newaz_uav-based_2016}; the localization is performed with a variational Bayesian algorithm. The authors of \cite{ristic_information_2010} propose 2D localization via a particle filter involving progressive correction and apply a search strategy based on maximizing the Rényi divergence; a relevant experimental verification demonstrated the capability of retrieving up to two sources. In article \cite{anderson_mobile_2022}, a particle filter is also used to localize sources in 3D; moreover, the radioactive decay and attenuation of the radiation in the obstacles can be modeled, thanks to automatic identification of the isotopes. During the search, a pre-determined number of measurements are conducted; the choice of the optimal trajectory is FIM-based. Nonetheless, the above-mentioned studies do not demonstrate the capability of retrieving multiple sources while ensuring that the entire region is explored; importantly, our research proposes an attempt to address such a deficiency.

The problems, scenarios, and preconditions in this article can be described as follows: Let us have an unknown number of radioactive sources that are concealed in a known region of interest (ROI) defined by a polygon with holes (static obstacles). A single unmanned ground vehicle (UGV) equipped to control its linear and angular velocity is available; the UGV carries an accurate self-localization system and a radiation detector which provides a counts per second (CPS) value at a constant sampling period. The goal is to localize all of the sources as quickly as possible; the result is expected to be independent of the starting position of the robot, and the robot must not leave the ROI or cross the obstacles.
The proposed algorithms are verified through simulations exploiting the real-world dataset acquired during our previous research \cite{gabrlik_automated_2021}.

\section{Localization algorithm}

This section characterizes the proposed localization algorithm, which exploits the importance sampling principle. The method has been designed to function independently of the data acquisition trajectory, and it should operate smoothly even in pre-planned systematic surveys. The particle filter is a Bayesian technique that approximates a posterior distribution by a set of random samples, i.e., particles \cite{thrun_probabilistic_2005}. Let us have a state vector $\bm{\theta}$ and a set of observations $\mathbb{Z} = 
\{ \bm{z}_i \}_{i=1}^M$
. At a time step $t$, the posterior probability is computed via the Bayes rule

\begin{equation}
	p(\bm{\theta}_t | \bm{z}_t) = \frac{p(\bm{z}_t | \bm{\theta}_t) \cdot p(\bm{\theta}_t | \bm{z}_{1:t-1})}{p(\bm{z}_t | \bm{z}_{1:t-1})}.
\end{equation}

As the number of sources $r$ is unknown, it embodies one of the estimated state variables, and the length of the vector $\bm{\theta}$ varies accordingly. Let us have a set of $N$ particles $\bm{\chi}_t = 
\{ \bm{\theta}^{(i)}_t \}_{i=1}^N$, where $\bm{\theta}^{(i)} = (r, \lambda_\mathrm{B}, x_1, y_1, \lambda_1, \cdots, x_r, y_r, \lambda_r$). The mean background radiation rate is denoted as $\lambda_\mathrm{B}$, and the tuple $(x,y,\lambda)$ represents the 2D coordinates of the source, together with its mean count rate at the distance of one meter.

At the start of the localization process, the particles are initialized randomly. We then have to select the maximum number of sources, $r_\mathrm{max}$, with the minimum assumed to equal one. Although an emitter is assumed to be present during the initialization phase, the algorithm is capable of exploiting a particle regularization to reach the hypothesis that there are no sources (see below). The prior probability of $r$ sources being present is adopted from \cite{ristic_information_2010}; this probability drops linearly with the increasing $r$. The background radiation, $\lambda_\mathrm{B}$, is distributed uniformly. The sources' coordinates $(x,y)$ are drawn from the uniform distribution, and samples outside the outer boundaries $\mathcal{R}$ of the ROI are rejected. Finally, the intensity $\lambda$ follows the gamma distribution $\Gamma(\alpha, \beta)$, the two parameters being the shape and the rate, respectively. 

Traditionally, the particle filter involves a prediction step that reflects the state transition probability $p(\bm{\theta}_t | \bm{\theta}_{t-1}, \bm{u}_t)$ given by the previous state and the control input $\bm{u}_t$. In this case, we assume the system to be stationary, i.e., $\bm{\theta}_t = \bm{\theta}_{t-1}$. Such simplification is possible with the sources in static positions and their half-life values markedly exceeding the duration of the localization; thus, the radioactive decay can be ignored.

A correction step follows, each particle being assigned a weight computed according to the measurement model 
\begin{equation}
	w^{(i)}_t \propto p(\bm{z}_t | \bm{\theta}^{(i)}_t) \cdot w^{(i)}_{t-1}.
\end{equation}
To derive a suitable model, four effects have to be considered:

1. Both the radioactive decay and the radiation detection are stochastic processes, meaning that we need to select an appropriate probability density function (PDF) to represent adequately the relevant physical laws. The radioactive decay follows a binominal distribution, commonly replaced with a Poisson distribution having a mean $\lambda$ \cite{foster_binomial_1983}. At large rates, we can further apply an approximation by the normal distribution whose mean and variance equal $\lambda$, that is

\begin{equation}
	\mathcal{P}(\lambda) \sim \mathcal{N}(\lambda, \lambda),
\end{equation}
\begin{equation}
	p(X=k) = \frac{\lambda^k e^{-\lambda}}{k!} \approx \frac{1}{\sqrt{\lambda 2 \pi}} e^{-\frac{(k-\lambda)^2}{2\lambda}}.
\end{equation}

2. Gamma radiation propagates with respect to the inverse square law and is attenuated by the mass it passes through. Ideally, with no scattering and secondary radiation, the intensity decreases to
\begin{equation}
\label{eq:propagation}
	I=I_0 \frac{\exp \left( -\sum_{i=1}^{S} \mu_i d_i \right)}{\left( \sum_{i=1}^{S} d_i \right) ^2},
\end{equation}
where $I_0$ is the initial intensity, $\mu_i$ represents the linear attenuation coefficient of the $i$-th material, and $d_i$ is the length of the intersection of a hypothetical radiation ray with the material \cite{knoll_radiation_2010}. Our scenario considers high-energy photons passing only through air at relatively short distances (< 20 m); therefore, the attenuation effect can be ignored, and the numerator expression in Eq.~\ref{eq:propagation} approximately equals one. Conversely, reflecting the attenuation would significantly increase the complexity of the estimation problem, as the attenuation coefficient $\mu$ is energy-dependent; thus, we cannot know its value a priori.

3. The radiation background introduces considerable noise into the measurements. The relevant components include terrestrial radiation, produced by the radionuclides that are naturally present in the soil, and galactic and solar cosmic radiation. It has to be considered that each acquired spectrum or count rate embodies a superposition of the useful signal yielded by the localized sources on the one hand and the background on the other \cite{shahbazi-gahrouei_review_2013}.

4. The detection system may suffer from dead time when overloaded with a high flux of photons. In particular situations, above all, those where the system exhibits paralyzable behavior, the detected counts start to decrease with increasing actual photon interactions; such an effect occurs when the rates are high. This condition can be compensated for by computing the expected counts $\lambda'$, using the theoretical rate $\lambda$ and the detector-specific dead time constant $\tau$ \cite{usman_radiation_2018}.

Combining all of the above effects enables us to express the probability $p(\bm{z}_t | \bm{\theta}^{(i)}_t)$. The measurement vector $\bm{z}_t$ is characterized by the tuple $(\phi_t, \psi_t, \nu_t)$, that is, the coordinates in the $x$ and $y$ axes, and the detected count rate. First, we need to compute the theoretical count rate at the point $(\phi_t, \psi_t)$, yielded by $\bm{\theta}^{(i)}_t$:
\begin{equation}
\label{eq:countrate}
	\lambda(\bm{z}_t,\bm{\theta}^{(i)}_t) = \lambda^{(i)}_\mathrm{B} + \sum_{j=1}^{r^{(i)}} \frac{\lambda^{(i)}_j}{(x_j^{(i)}-\phi_t)^2 + (y_j^{(i)}-\psi_t)^2 + D^2},
\end{equation}
where $D$ is the fixed detector height above the ground level; note that we anticipate all of the sources to be located on the ground. Subsequently, the dead time effect is applied:
\begin{equation}
\label{eq:deadtime}
	\lambda' = \lambda(\bm{z}_t,\bm{\theta}^{(i)}_t)\cdot e^{-\tau\cdot\lambda(\bm{z}_t,\bm{\theta}^{(i)}_t)}.
\end{equation} 

The notation has been slightly simplified, yielding the reduced equation below, which computes the unnormalized weight:
\begin{equation}
	\hat{w}_t^{(i)} = \frac{1}{\sqrt{\kappa\lambda' 2 \pi}} e^{-\frac{(\nu_t-\lambda')^2}{2\kappa\lambda'}} \cdot w_{t-1}^{(i)},
\end{equation}
where $\kappa$ is the constant that helps us to tune the variance of the utilized normal PDF to respect the real-world measurements. Once the particles have been processed, the weights are normalized, and the effective sample size $N_\mathrm{eff}$ is computed. We have

\begin{equation}
	w_t^{(i)} = \frac{\hat{w}_t^{(i)}}{\sum_{j=1}^N \hat{w}_t^{(j)}}; \hspace{0.5cm}
	N_\mathrm{eff} = \frac{1}{\sum_{i=1}^N (w_t^{(i)})^2}. 
\end{equation}

To prevent particle depletion, resampling is not performed in each iteration; instead, the algorithm idles until the effective sample size has dropped below the chosen threshold, $N_\mathrm{eff} < N_\mathrm{thr}$. Eventually, the resampling is executed using the low variance algorithm \cite{hol_resampling_2006}.

As each resampling operation reduces the particle set variance, this needs to be increased via regularization. Such a step also helps the localization algorithm to respond to newly discovered sources through altering their estimated count $r$. First, all parameters but $r$ are regularized. The resampled set $\bm{\chi}$ is divided into subsets, $\bm{\chi}_1, \bm{\chi}_2, \cdots, \bm{\chi}_{r_\mathrm{max}}$, with respect to the number of sources. In each subset, the standard deviation $\bm{\sigma}$ of the parameters is computed, and the vector of random numbers $\bm{G}$ is drawn from the Gaussian kernel. Then, the particles are updated to read

\begin{equation}
	\forall \bm{\theta} \in \bar{\bm{\chi}}_i: \bm{\theta} = \bm{\theta} + \frac{h_i}{\xi}\bm{\sigma}_i \bm{G},
\end{equation}
where $h$ is the suggested bandwidth \cite{musso_improving_2001}, and $\xi$ denotes the tuning parameter. Whenever a source hypothesis reaches beyond $\mathcal{R}$ or its intensity drops below zero, it dissolves, and the respective $r$ value is decremented. 

Then, the number of sources is regularized according to the pre-set probabilities of 'birth', $p_\mathrm{B}$, and 'death,' $p_\mathrm{D}$. The latter case is straightforward: a random source hypothesis is picked and dissolved; note that the minimum allowed number of sources equals zero. When a new hypothesis is added, the corresponding parameters are sampled, respecting the posterior $p(\bm{\theta}_t | \bm{z}_t)$. Specifically, the coordinates $(x,y) \in \mathcal{R}$ are sampled from the normal distribution centered at $(\phi_t, \psi_t)$, while the intensity exploits
\begin{equation}
	\lambda \sim \mathcal{P}\left(\nu_t\left[(x-\phi_t)^2 + (y- \psi_t)^2 + D^2\right] \right).
\end{equation}

The control algorithm presented in the next section requires us to use the current source estimate in some cases; to acquire one, the expected number of sources is computed first, reading
\begin{equation}
	\hat{r}_t = \Biggl\lfloor \sum_{i=1}^N w_t^{(i)}\cdot r_t^{(i)} + 0.5\Biggr\rfloor.
\end{equation}

The source estimate is then expressed as
\begin{equation}
	\hat{\bm{\theta}}_t = \left. \sum_{i: r^{(i)}=\hat{r}_t} w_t^{(i)} \cdot \bm{\theta}^{(i)} \middle/ \sum_{i: r^{(i)}=\hat{r}_t} w_t^{(i)}. \right. 
\end{equation}

\section{Control algorithm}

The control system is designed to propose a motion command on the basis of previous measurements. The suggested algorithm comprises two components, one denoted as a \emph{local} and the other as a \emph{global} planner; the former aims to speed up the convergence of the particle filter, while the latter ensures that the whole ROI is covered. 

The region is subjected to an approximate cell decomposition, which yields a set of free square cells $\mathbb{C} = \{\mathcal{C}_i\}_{i=1}^K$. The cells have a dimension $\delta$, chosen with respect to the time efficiency relative to the exhaustive exploration along a boustrophedon path. The extent of the most dense meaningful trajectory can be pre-determined for a given ROI by the parameters of the detection system and the desired spatial resolution; this limit should not be exceeded in the dynamic planning.
The cell $\mathcal{C}_i$ is considered free when its center, $\bm{c}_i$, lies inside the ROI, $\mathcal{R}$, and does not appear within any obstacle $\mathcal{O}\in\mathbb{O}$.
A survey ends once every cell has been \emph{visited}, meaning that the number of measurements acquired therein is greater than the preset threshold:
\begin{equation}
	\forall \mathcal{C} \in \mathbb{C}:\hspace{0.3cm} \big| \{ {\bm{z}_i: (\phi_i, \psi_i) \in \mathcal{C}} \} \big| \ge s_{\mathrm{min}}.
\end{equation}

The two planners introduced earlier are switched according to three conditions.
In these, the global planner is applied if: 1. The current cell (i.e., that which accommodates the robot at a time $t$) is visited; 2. the robot is not in a free cell (as may happen near region boundaries and obstacles); 3. the relative mean unnormalized weight of the particles is above the threshold
\begin{equation}
	\overline{\hat{w}_t} \hspace{0.1cm} / \hspace{0.1cm} \overline{\hat{w}_{1:t}} > \hat{w}_\mathrm{thr}.
\end{equation}
The local planner, by contrast, finds use in all other scenarios.
Both planners differ in the criterion function $f(\bm{u})$, which allows selecting the fittest member from the set of potential actions $\mathbb{U} = \big\{ \bm{u}_i=(v_i, \omega_i) \big\}_{i=1}^L $ (the linear and angular velocities). In each action, a new position $(\phi', \psi')$ is acquired with a common differential drive kinematic model. % \cite{hellstrom_kinematics_2011}.

The local planner relies on the Shannon entropy. First, the expected count rate at $(\phi', \psi')$ is computed with respect to the current estimate $\hat{\bm{\theta}}_t$ via Eqs.~\ref{eq:countrate} and \ref{eq:deadtime}; subsequently, all particles are weighted. The entropy is then given by
\begin{equation}
	H= - \sum_{i=1}^N w^{(i)}\cdot\log w^{(i)}.
\end{equation}
Finally, the entropy values are rescaled so that the maximum equals one. 

Conversely, the global planner's criterion exploits the Euclidean distance between the new position and the center of the next-best cell. To choose this cell, the \emph{curiosity} value $C$ is estimated, equaling the standard deviation of expected measurements at the center $\bm{c}_i$ of a cell, the measurements being yielded by the particles in $\bm{\chi}_t$. We have
\begin{equation}
	C(\mathcal{C}_i) = \sqrt{\frac{1}{N}\sum_{j=1}^N  \left( \mathbb{E}[\lambda(\bm{\theta}^{(j)}, \bm{c}_i)] - \overline{\mathbb{E}[\lambda}] \right)^2}.
\end{equation} 

As the most curious cell may appear on the opposite side of the ROI and the curiosity may change with every measurement, an A*-like algorithm is adopted to pick a suitable unvisited cell that is close to the robot. This algorithm searches for an optimal path from the current position to the highest-curiosity cell; however, the cost of visiting a node (cell) is not only given by the physical distance but also exhibits an inverse proportionality to the respective curiosity value. The resulting next-best cell then embodies the first unvisited node along the path. Once this sub-goal has been reached, a new one is chosen.

At the following stage, obstacles need to be considered, as we do not desire the robot to cross them. To address this requirement, another criterion function common for the planners is introduced; the function is inspired by the repulsive force used in artificial potential fields \cite{rostami_obstacle_2019}, reading
\begin{equation}
\begin{aligned}
	g(\bm{u}) = & \left(R_\mathrm{R} - \min||(\phi', \psi') - (x,y)\in \mathcal{R}|| \right)^2 + \\
	& \sum_{\mathcal{O}\in\mathbb{O}} \left(R_\mathrm{O} - \min||(\phi', \psi') - (x,y)\in \mathcal{O}|| \right)^2,
\end{aligned}
\end{equation}
where $R_\mathrm{R}$ and $R_\mathrm{O}$ represent the effective radii for the region boundaries and the obstacles, respectively. Finally, the fittest action is selected by using the criterion functions (depending on the currently applied planner) and weighting factors $a, b$: 
\begin{equation}
	\underset{\bm{u}\in\mathbb{U}}{\arg\min} \hspace{0.3cm} a \cdot f(\bm{u}) + b\cdot g(\bm{u}).
\end{equation}

\begin{figure}[thpb]
	\centering
	\includegraphics[width=0.36\textwidth]{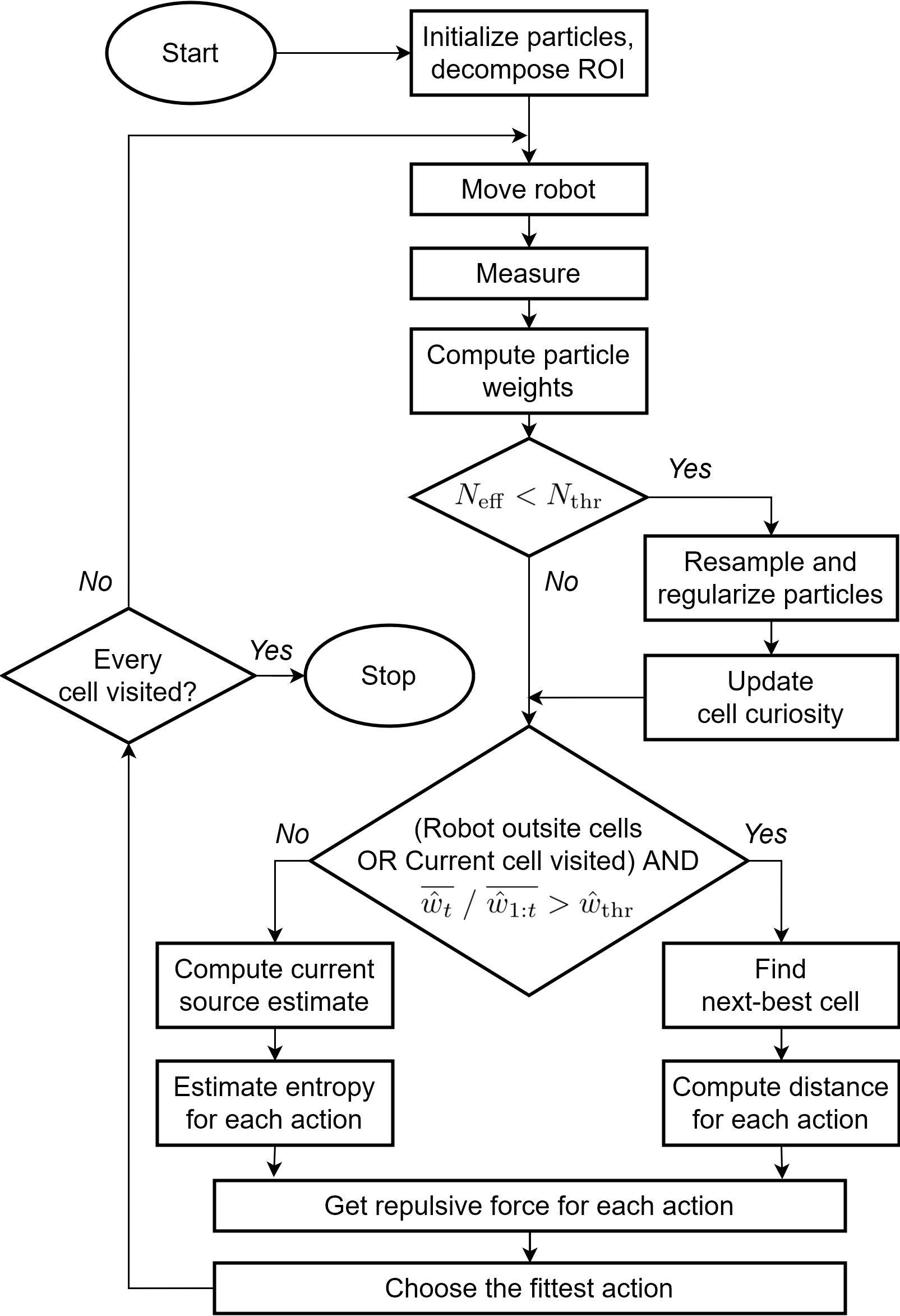}
	\caption{The structure of the localization and control algorithms.}
	\label{fig:flowchart}
\end{figure}

The overall structure of the proposed algorithm is outlined through the flowchart in Fig.~\ref{fig:flowchart}.

\section{Experimental setup}

The performance of both the localization and the control algorithms was tested via a comprehensive simulation study utilizing real-world experimental data \cite{gabrlik_automated_2021}. The region of interest captures an area of 277 m$^2$ and contains three static obstacles, whose areas range from 1 m$^2$ to 2.5 m$^2$  and were delimited utilizing a photogrammetric model. The obstacles in this scenario involved exclusively sparse vegetation and plastic drums, allowing them to avoid conflict with ignoring the attenuation. It is assumed that the survey can start in any vertex of the polygon $\mathcal{R}$. The ROI comprises three distinguishable radiation sources, of which one is cesium-137 and two are cobalt-60, the respective activities calculated to the measurement date being 80~MBq, 25~MBq, and 3~MBq; hereafter, the point sources are specified as $S_1$, $S_2$, and $S_3$. Note that the radioactive material is sealed and unshielded.

The area was decomposed to 29 cells, each having the dimension $\delta=3$ m (Fig.~\ref{fig:celldecomp}). The character of the study site, i.e., a flat grass field, enables simplifying the localization algorithm into two dimensions. We can reasonably assume that any uncontrolled point source lies on the terrain surface, which is known to be flat thanks to the available digital elevation model (DEM).

The original data were acquired by an Orpheus-X4 UGV carrying a pair of $2^{\prime\prime} \times 2^{\prime\prime}$ thallium-doped sodium iodide (NaI(Tl)) detectors that executed the sampling at the period of 1 s. The self-localization was ensured by an accurate Real-time Kinematic Global Navigation Satellite System (RTK-GNSS) receiver. With respect to the applied platform's capabilities and limitations, the set of candidate actions was populated with 5 elements: $\mathbb{U}=\{ (0.6$ ms$^{-1}, 0$ s$^{-1}), (0.5$ ms$^{-1}, \pm \pi/8$ s$^{-1}), (0.4$ ms$^{-1}, \pm \pi/4$ s$^{-1}) \}$.
The remaining relevant parameters are specified in Table~\ref{tab:parameters}. 

\begin{table}[h]
	\caption{The relevant parameters of the proposed algorithms.}
	\label{tab:parameters}
	\begin{center}
		\begin{tabular}{cc|cc|cc|cc}
			\hline
			$N$ & $10^4$ & $\alpha$ & $2$ & $N_\mathrm{thr}$ & $2000$ & $s_\mathrm{min}$ & $3$ \\
			
			$r_\mathrm{max}$ & $10$ & $\beta$ & $12000$ & $\xi$ & $2.55$ & $\hat{w}_\mathrm{thr}$ & $0.35$ \\
			
			$\lambda_\mathrm{B,min}$ & $250$ & $\tau$ & $2\cdot10^{-5}$ & $p_\mathrm{B}$ & $1 / 100$ & $R_\mathrm{R}$ & $1.5$ m \\
			
			$\lambda_\mathrm{B,max}$ & $750$ & $\kappa$ & $15^2$ & $p_\mathrm{D}$ & $1 / 600$ & $R_\mathrm{O}$ & $1$ m \\
			\hline
		\end{tabular}
	\end{center}
\end{table}

\section{Results and Discussion}

\begin{figure}[thpb]
	\centering
	\includegraphics[width=0.35\textwidth]{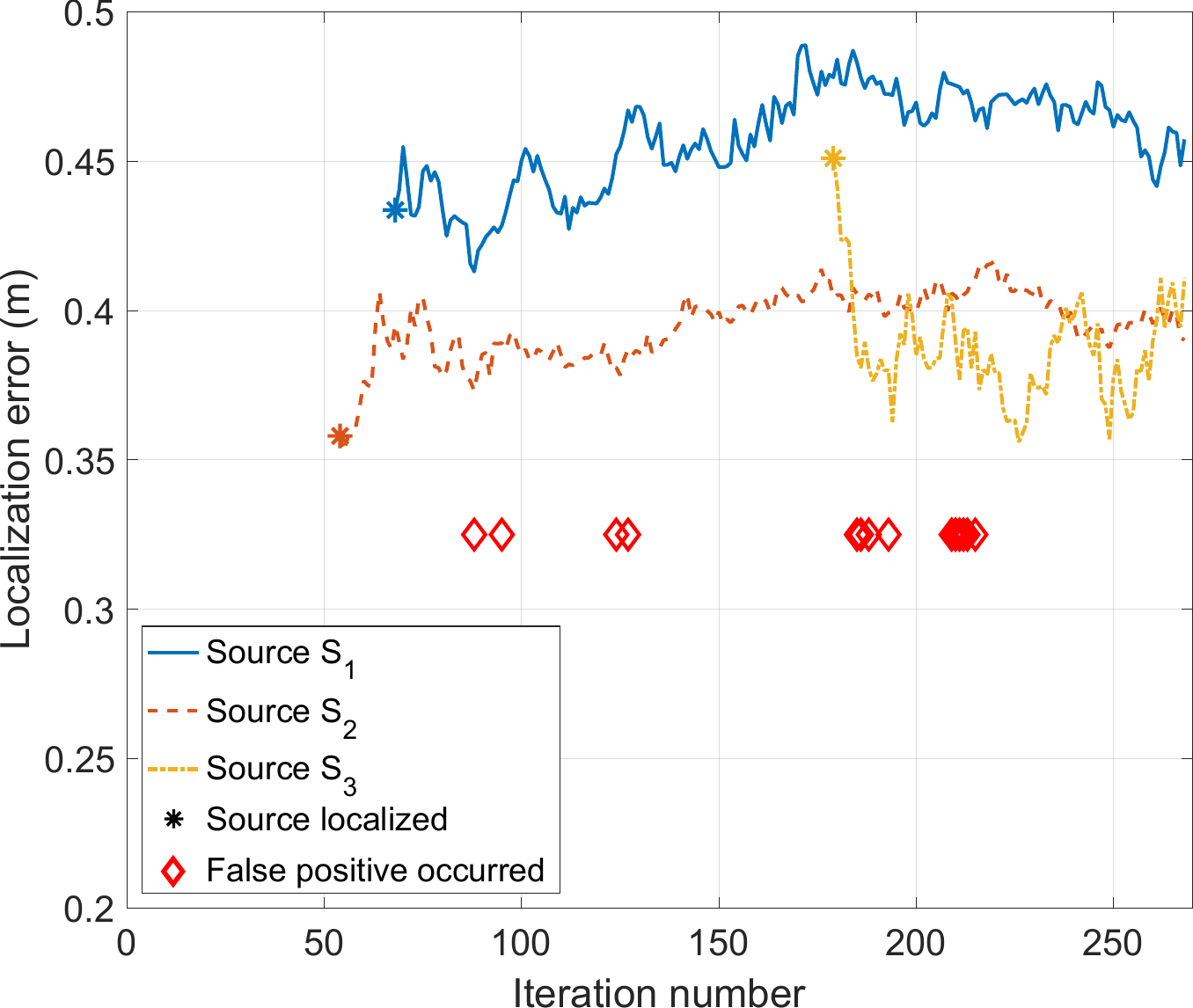}
	\caption{The localization error pattern.}
	\label{fig:locerr}
\end{figure}

\begin{figure*}[thpb]
	\begin{subfigure}{0.45\textwidth}
		\centering
		\includegraphics[width=.8\linewidth]{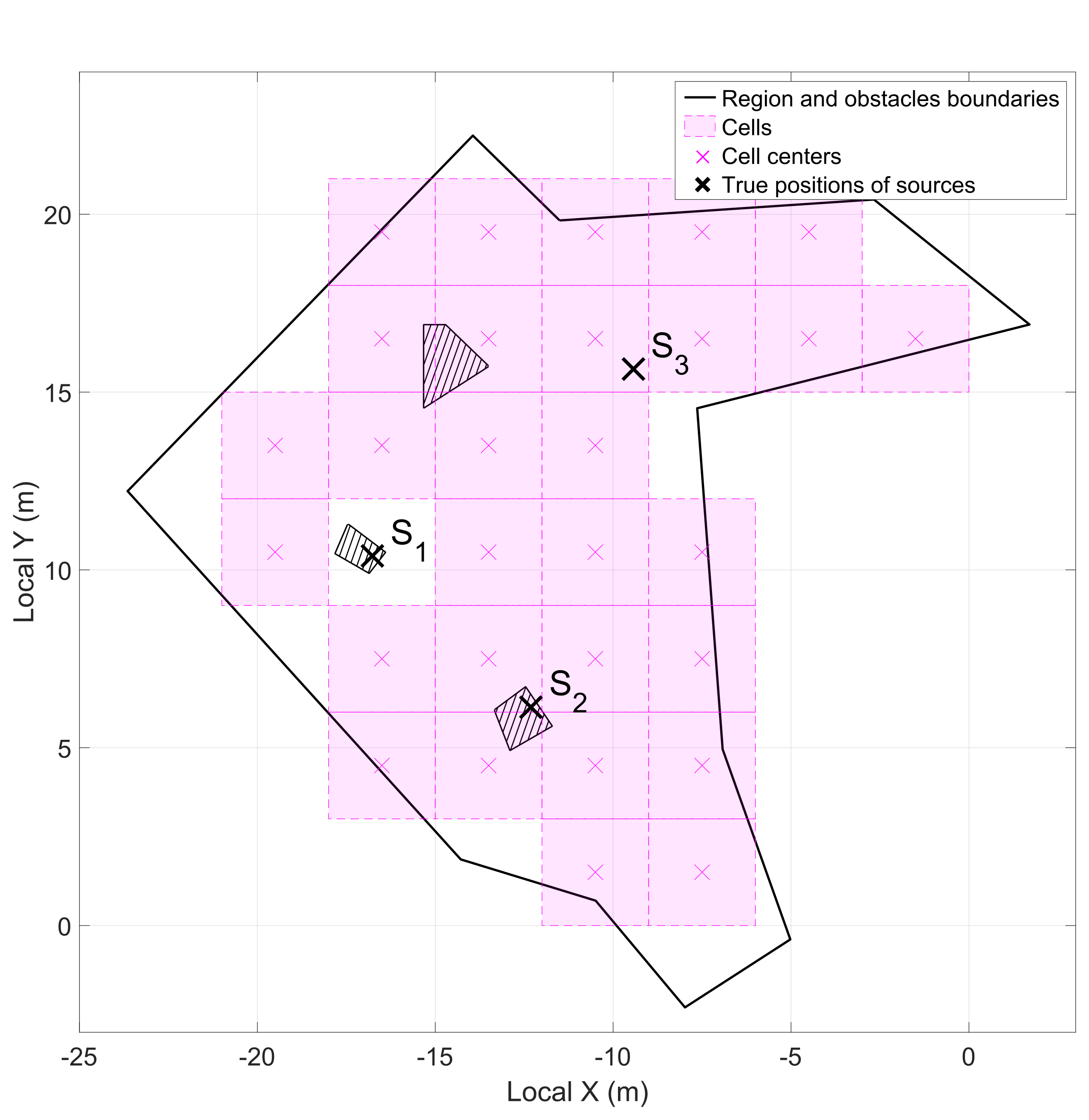}  
		\caption{}
		\label{fig:celldecomp}
	\end{subfigure}
	\begin{subfigure}{0.45\textwidth}
		\centering
		\includegraphics[width=.86\linewidth]{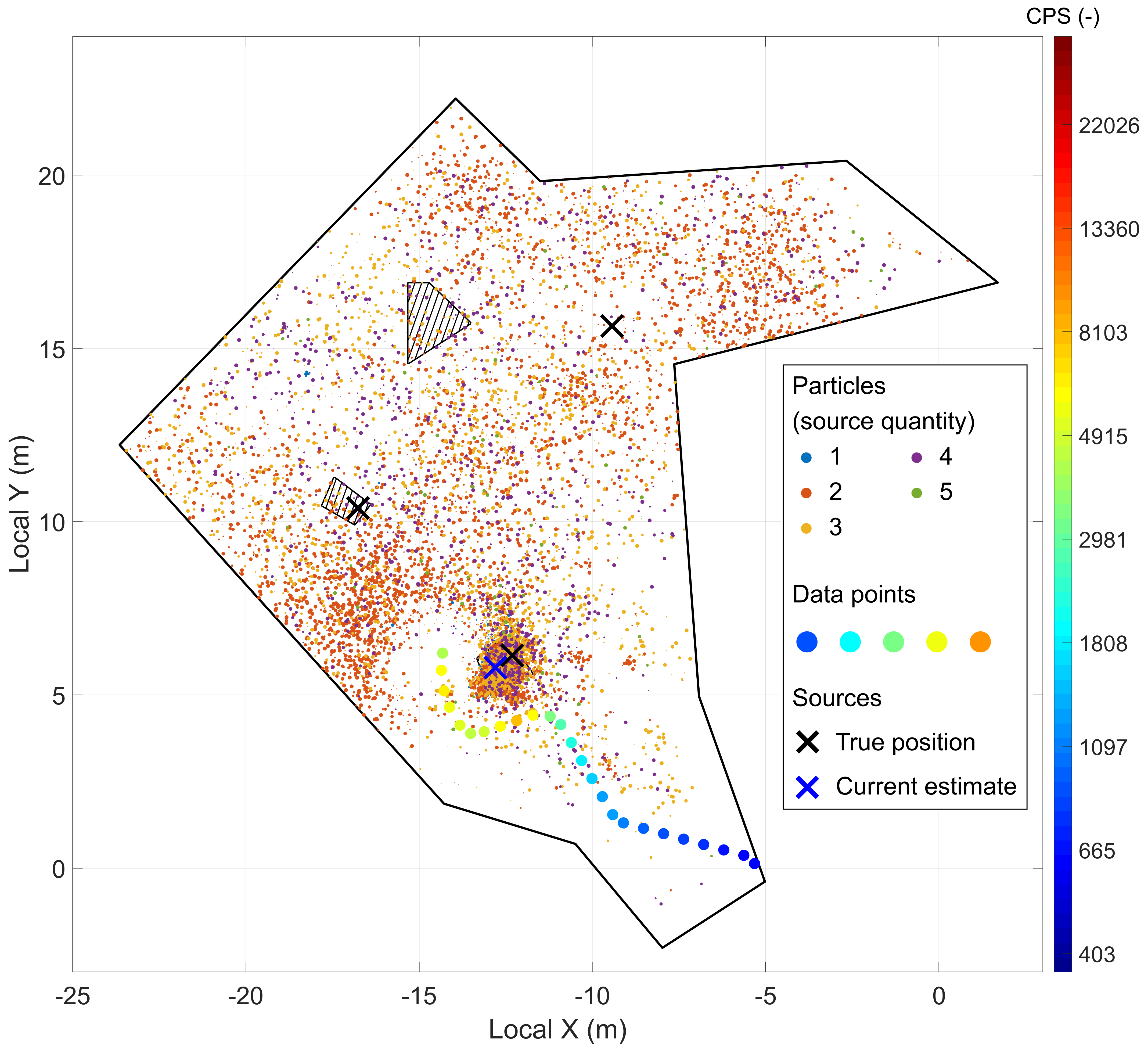}  
		\caption{}
		\label{fig:iteration1}
	\end{subfigure}
	\newline
	\begin{subfigure}{0.45\textwidth}
		\centering
		\includegraphics[width=.86\linewidth]{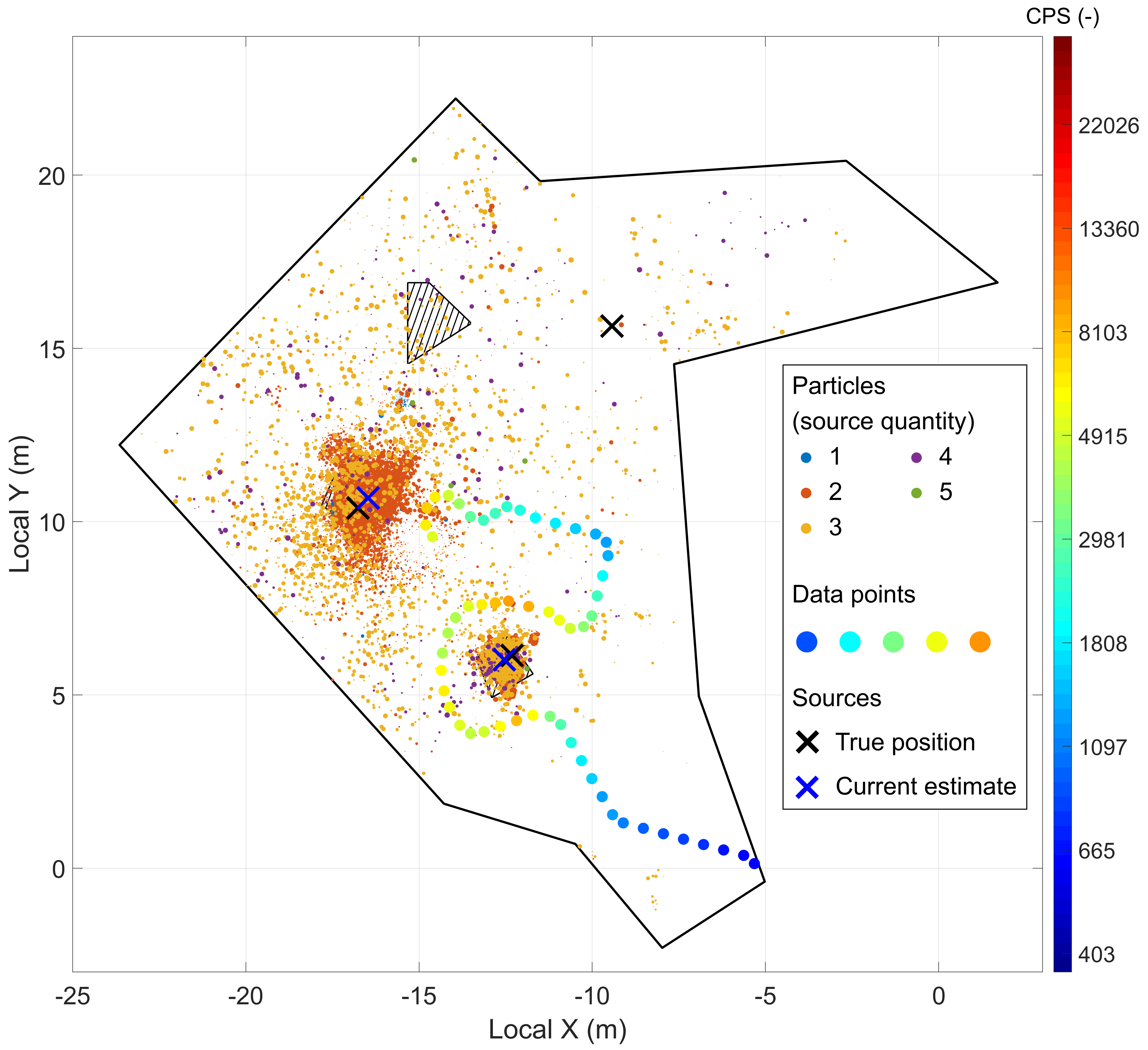}  
		\caption{}
		\label{fig:iteration2}
	\end{subfigure}
	\begin{subfigure}{0.45\textwidth}
		\centering
		\includegraphics[width=.86\linewidth]{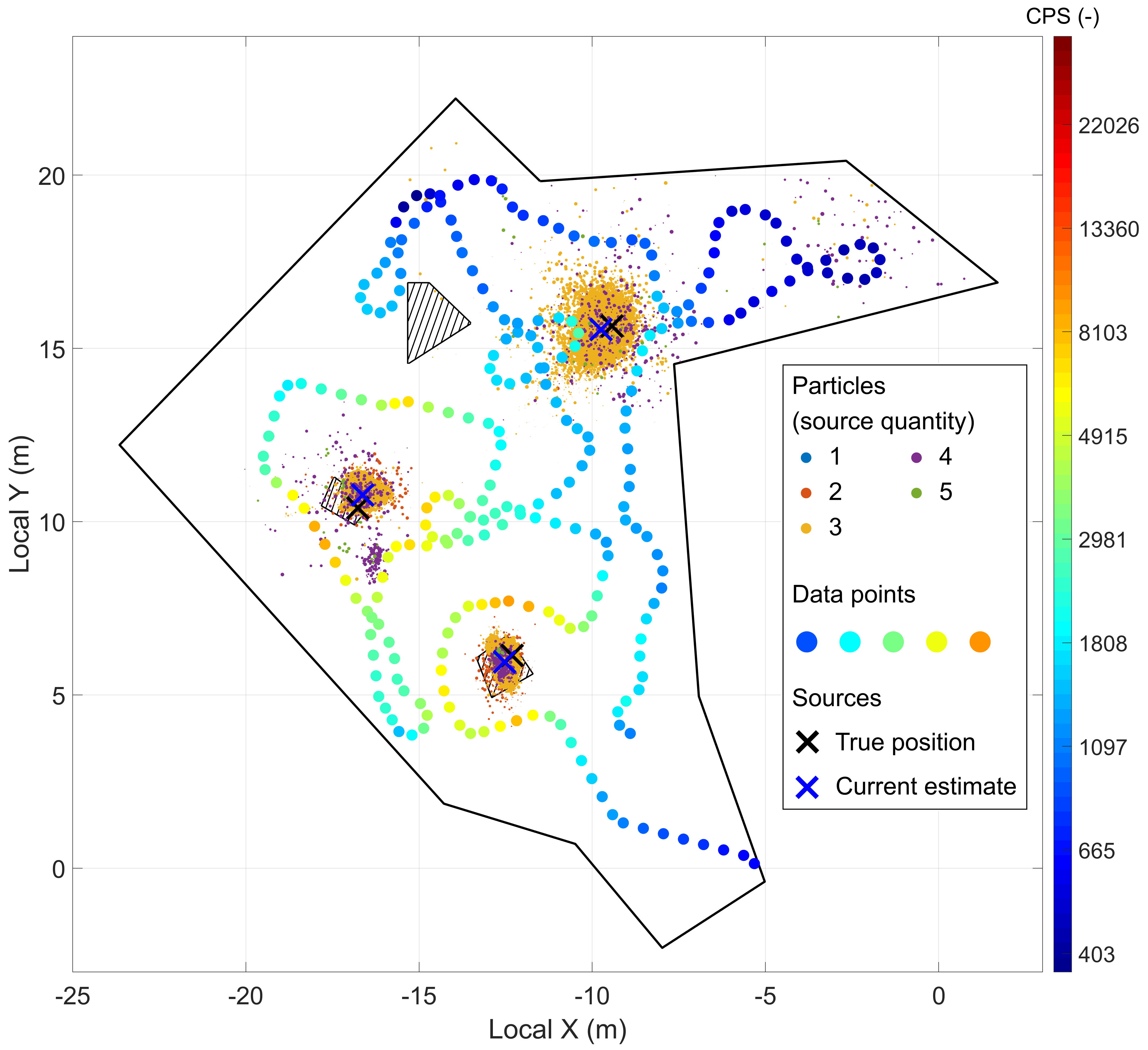}  
		\caption{}
		\label{fig:iteration3}
	\end{subfigure}
	
	\caption{The region of interest decomposed into cells, with the sources denoted alphanumerically (a). Three iterations of an example run of the proposed algorithm: Showing iterations No. 25 (b), 56 (c), and 229, namely, the final one (d). The color bars are in the log scale.}
	\label{fig:mainimages}
\end{figure*}

Three iterations of an example run of the proposed algorithm are presented in Fig.~\ref{fig:iteration1}, \ref{fig:iteration2}, and \ref{fig:iteration3}, respectively. In this case, the survey took 229 iterations in total, and all of the three sources were localized successfully. To assess the efficiency of the control algorithm and the robustness of the localization one, 500 simulations were run, with the initial robot position being randomly selected from the set of the ROI vertices. As a reference, the dataset acquired during the pre-planned survey was employed; this dataset consisted of 437 datapoints iteratively fed into the localization algorithm. We carried out 100 simulations for both the original and the reversed measurement orders. An overview of the results is provided in Table~\ref{tab:results}; here, a source hypothesis is considered \emph{valid} once the variance of coordinates in both axes has dropped below 1.5~m$^2$, and a source is \emph{localized} if the corresponding hypothesis lies within a range of 1.5~meters. Valid hypotheses beyond 1.5~m from any source are labeled as \emph{false positives}. The progress of the localization error and the occurrences of the false positives in time are displayed in Fig.~\ref{fig:locerr}; note that these aspects embody the averaged results from all of the 500 simulations covered by our control algorithm. An additional series of simulations enabled us to verify how the algorithm performs in the no-source scenario; relevant results are summarized at the bottom of Table~\ref{tab:results}. The actual radiation measurements in this case were replaced with Gaussian noise having parameters ($\mu=500,\sigma=70$) derived from real data.

Considering the above-presented data, the localization algorithm can be characterized as robust enough, as it identifies all of the sources in each of the cases under the pre-planned exhaustive survey trajectory scenario. Such a good result, however, does not apply to source $S_3$ when the proposed (\emph{dynamic}) control algorithm is employed: The source is really weak and thus detectable only from a close proximity (< 1.5 m), and, given the selected cell size, the algorithm does not always navigate the robot adequately. As determined empirically, the current algorithm setup requires approximately 9 samples per cell on average, meaning that the $\delta$ value cannot be significantly reduced; for the studied area and the reference systematic survey, the information-driven control tends to be less efficient if $\delta < 2.3$ m. Note that the original line spacing equaled approximately 1~m, this being the value chosen to achieve the maximum possible spatial resolution with respect to the applied platform and detectors.

Interestingly, the false positive (FP) rate is significantly greater when the order of the datapoints has been reversed in the pre-planned trajectory case; this condition may arise from the fact that essentially all of the measurements in the first third of the survey carry only a minor information value, as the count rates are situated near the radiation background. Conversely, the dynamic planner exhibits a satisfactory FP rate, one that is comparable to the result achieved with the original dataset. In Fig.~\ref{fig:locerr}, the FPs are shown to appear mostly after the source $S_3$ has been encountered, with the other two sources having been already localized by that moment. Even though the algorithm may seem to encounter issues at low count rates due to an improper choice of the kernel for the particle weighting process, other functions were rejected, as they caused fast particle deprivation and overall algorithm instability: For example, the apparently suitable Poisson kernel exhibited an excessively narrow PDF in the given context. The algorithm's performance at low intensities was partially improved by progressively altering the $\kappa$ parameter (instead of leaving it constant); this approach, however, produced additional issues. The problem therefore needs to be addressed in the future research to yield more sophisticated adjustment of the method; possibly, some factors such as the directional characteristics of the detection system should not be neglected.

The localization error ranges from 0.35~m to 0.5~m and is relatively stable in each source during the experiment.
Such an accuracy may suffice from a practical perspective, but when really necessary, better results are achievable via post-processing, by using, for instance, the Gauss-Newton method \cite{gabrlik_automated_2021}. Moreover, it was demonstrated that an absence of sources does not affect the algorithm's behavior negatively: No FPs occurred during the surveys, and the estimated number of sources converged towards zero in both the pre-planned and the dynamic trajectories.

The proposed dynamic control has met our expectations, as it indeed reduces the time required to localize the sources independently of the robot's starting position; this holds true especially of the two strong sources, $S_1$ and $S_2$. The total iterations are reduced by 39 \% compared to the systematic approach, and the iterations needed to localize the three sources drop by 41 or 44 \%, depending on the order of the datapoints in the reference survey. The main drawback lies in that the weakest source, $S_3$, is not found each time. This issue is planned to be addressed in the future experiments, by such means as enhancing the planners to reward the actions which bring the robot farther from the previously acquired datapoints; this concrete step will increase the effective coverage of the ROI for the same number of iterations. 

Our current efforts were inspired mostly by Ristic et al. \cite{ristic_information_2010} and Mascarich et al. \cite{mascarich_autonomous_2022}. From the former, we adopted the regularization framework and particle structure, albeit with a slight modification: We allowed also the estimation of the mean radiation background rate (as suggested in, e.g., \cite{tan_fast_2022}). The latter then led us to develop the idea of dividing the control algorithm into local and global components. Our local planner exploited the Shannon entropy (\cite{huo_autonomous_2020}, \cite{pinkam_informative_2020}); although we had already carried out experiments involving the FIM, applying the entropy enabled us to obtain better results. 

The novelty of the research presented herein rests in the global planner design and the strategy of switching the control modes. In this context, we also modified the algorithms and tuned their parameters to reach sufficiently consistent outcomes even with noisy real-world data.  
Compared to Ristic et al., we demonstrated the ability to acquire valid localization results even when the parameter $r_\mathrm{max}$ is significantly greater than the actual number of sources being sought; moreover, our approach ensures complete coverage of the target area. 

In the future experimentation, we will focus on utilizing the information embedded in the measured radiation spectra; an inspiring option was presented by, for instance, Anderson et al. \cite{anderson_mobile_2022}. Regrettably, relevant datasets available to us lack reliable spectra because the detection system was damaged during the initial fieldwork. Another challenge to improve the procedures lies in exploiting the partial directional information provided by an array of measurement units; importantly, the task may be successfully completed with only two detectors. By extension, we can also mention that the presented use case including only bare sources may not be realistic; our planned work is therefore expected to focus on more complex scenarios.

\begin{table*}[thpb]
	\caption{The averaged localization results: The numbers after the $\pm$ sign represent the standard deviation from all of the performed simulated experiments (where applicable).}
	\begin{center}
		{\def\arraystretch{1}
			\begin{tabular}{llccc}
				\hline
				& & \bf{Proposed control} & \bf{Systematic survey}  & \bf{Systematic survey} \\
				& & \bf{algorithm} & \bf{(original)} & \bf{(reversed)} \\
				\hline
				\multicolumn{2}{c}{Total iterations} & $271\pm 32$ & $437$ & $437$ \\
				\multicolumn{2}{c}{False positive rate (\%)} & $4.8$ & $1.8$ & $9.0$ \\
				\hline
				$S_1$ & Localized? (\% of all experiments) & $99.8$ & $100.0$ & $100.0$ \\
				& First localizing iteration  & $77\pm 38$ & $64\pm 5$ & $217\pm 31$ \\
				& \% of localizing iterations  & $58\pm 18$ & $75\pm 11$ & $34\pm 9$ \\
				& Localization error (m) & $0.45\pm 0.29$ & $0.33\pm 0.24$ & $0.46\pm 0.37$ \\
				\hline
				$S_2$ & Localized? (\% of all experiments) & $100.0$ & $100.0$ & $100.0$ \\
				& First localizing iteration  & $51\pm 45$ & $22\pm 3$ & $311\pm 35$ \\
				& \% of localizing iterations  & $74\pm 19$ & $76\pm 16$ & $26\pm 8$ \\
				& Localization error (m) & $0.39\pm 0.25$ & $0.43\pm 0.19$ & $0.48\pm 0.34$ \\
				\hline
				$S_3$ & Localized? (\% of all experiments) & $62.8$ & $99.0$ & $100.0$ \\
				& First localizing iteration  & $176\pm 65$ & $301\pm 11$ & $148\pm 4$ \\
				& \% of localizing iterations  & $24\pm 17$ & $30\pm 5$ & $29\pm 11$ \\
				& Localization error (m) & $0.47\pm 0.34$ & $0.38\pm 0.24$ & $0.33\pm 0.29$ \\
				\hline
				\hline
				\multicolumn{2}{l}{No sources} & & & \\
				& Total iterations  & $216\pm22$ & $437$ & $437$ \\
				& False positive rate (\%)  & $0.0$ & $0.0$ & $0.0$ \\
				& Estimated number of sources  & $1$e$-5 \pm3$e$-4$ & $0.012\pm0.005$ & $0.010\pm0.003$ \\
				\hline
		\end{tabular}}
	\end{center}
	\label{tab:results}
\end{table*}

\section{Conclusion}

The article discusses a comprehensive method for localizing multiple radiation sources by means of an autonomous mobile robot in a known outdoor environment. Two principal factors, namely, a localization and a control algorithm, are relied on: The former estimates the number of the sources and their relevant parameters via a particle filter, and the latter chooses the optimal robot movement sequence to reduce the time required to find the sources while ensuring complete coverage of the region of interest. The novelty of the research rests in conveniently combining known partial algorithms into a coherent unit that delivers robust performance, as verified through extensive simulation studies based on real-world data. Our solution localizes the sources already during the measurement, i.e., earlier than the post-processing, and alters the robot trajectory accordingly to prioritize the most information-rich sectors of the target area. The method is applicable primarily within the search for uncontrolled sources but can be modified to find use in other domains too. Importantly, the algorithms will be deployed on an Orpheus-X4 platform to yield reliable functionality verification.

The drawbacks include, above all, the need to know the environment map a priori, the dependence on an accurate RTK-GNSS self-localization, the assumption that the sources are located in a 2D plane, and the inability to manage the radiation attenuation in potentially dense obstacles. To address the first two issues, we intend to enable the UGV to navigate itself with lidar-based Simultaneous localization and mapping (SLAM); however, an instrument to limit the scope of the surveyed region will still be necessary. This planned step will improve the system's overall autonomy, albeit probably at the expense of the localization algorithm's performance, which may deteriorate due to a lower accuracy of the datapoint positioning. Regarding the third problem, the difficulty is easily resolvable through expanding the particle structure and the measurement model to include another coordinate; to avoid estimating phantom sources in improbable positions (e.g., hovering in air), it may be beneficial to acquire and exploit a terrain model.

The last of the above-outlined disadvantages, however, is markedly more prominent, requiring knowledge of the radiation energy, geometry of the obstacles, and relevant attenuation coefficients. A set of possible solutions were proposed in the literature; most of the authors nevertheless assume that at least some parameters have been provided in advance to reduce the estimation problem complexity. %Other suggestions for future work were offered in the previous section.

\bibliographystyle{unsrt}
\bibliography{references}

\end{document}